\documentclass[sigconf,nonacm]{acmart}
\usepackage{amsmath}
\usepackage{natbib} 
\usepackage{mathtools}
\usepackage[toc,page, title, titletoc]{appendix}
\usepackage{pgfplots}
\pgfplotsset{compat=newest}
\pgfplotsset{scaled y ticks=false}
\usepgfplotslibrary{groupplots}
\usepgfplotslibrary{dateplot}

\usepackage{tikz}
\usepackage{multirow}
\usepackage{makecell}

\AtBeginDocument{%
  \providecommand\BibTeX{{%
    \normalfont B\kern-0.5em{\scshape i\kern-0.25em b}\kern-0.8em\TeX}}}
    
\begin{document}

\title[]{Explainable Comparison of Feature-Based and Deep Learning Models for TROPOMI Methane Plume Screening}

\author[S.Kurchaba]{Solomiia Kurchaba}
\email{s.kurchaba@sron.nl}
\affiliation{
\institution{SRON Space Research Organisation Netherlands}
\city{Leiden}
\country{The Netherlands}
}

\author[Joannes D. Maasakkers]{Joannes D. Maasakkers}
\affiliation{
\institution{SRON Space Research Organisation Netherlands}
\city{Leiden}
\country{The Netherlands}
}

\author[Berend J. Schuit]{Berend J. Schuit}
\affiliation{
\institution{SRON Space Research Organisation Netherlands}
\city{Leiden}
\country{The Netherlands}
}
\affiliation{
\institution{GHGSat Inc.}
\city{Montreal}
\country{Canada}
}

\author[Ilse Aben]{Ilse Aben}
\affiliation{
\institution{SRON Space Research Organisation Netherlands}
\city{Leiden}
\country{The Netherlands}
}
\affiliation{
\institution{Department of Earth Sciences, Vrije Universiteit Amsterdam}
\city{Amsterdam}
\country{The Netherlands}
}

\date{May 2026}

\keywords{CH4, emissions, machine learning, methane, plumes, TROPOMI}
\begin{abstract}
 
Continuous and global detection of large methane emissions is a crucial step for global warming mitigation. Satellite observations, such as from S5P/TROPOMI, combined with plume detection algorithms, can play a key role in this effort. However, not all TROPOMI plume detections that look like methane emission plumes are the result of actual emissions. A significant part of the plume-like features in the data are retrieval artifacts. Such artifacts could be the result of variations in elevation or albedo gradients, high concentrations of aerosols, coastal lines, water bodies, etc. Previous work approached the problem of plume-artifact classification by means of a Support Vector Machine Classifier (SVC), trained on an extensive set of observation-based scalar features designed by the domain experts. However, such an approach limits the information scope received by the algorithm to what is deemed to be important by the experts, breaks the spatial relationship between pixels, and loses information during the process of statistical aggregation. In this study, we compare feature-based (SVC, Random Forest, XGBoost) and image-based (ResNet-18, ResNet-34) models for methane plume–artifact classification under balanced and imbalanced evaluation settings. To interpret the results, we apply SHAP-based explainability to both model families. Our findings provide practical guidance for model selection in operational methane-screening workflows such as the CAMS Methane Hotspot Explorer.

\end{abstract}
\maketitle

\section{Introduction}
\label{sec:introduction}
Anthropogenic methane emissions are responsible for more than 30\% of human-caused global warming \cite{ocko2018rapid, masson2021climate}. Moreover, a substantial fraction of total methane emissions originates from a relatively small number of very large sources, often referred to as “super-emitters”, including coal mines, oil and gas infrastructure, and landfills. Consequently, effective mitigation of global warming in the short term benefits from the continuous monitoring of these super-emitters. The TROPOspheric Monitoring Instrument (TROPOMI) on board the ESA Sentinel-5 Precursor (S5P) satellite \cite{veefkind2012tropomi} plays a key role in this monitoring. Launched in 2017, TROPOMI is the first instrument capable of detecting methane emission plumes from super-emitters globally on a daily basis \cite{jacob2022quantifying, acp-23-9071-2023}. Here, we compare feature-based and image-based machine learning techniques to screen plumes detected in these data for artifacts.

In satellite observations, emissions from super-emitters typically manifest as localized methane plumes, making plume detection in TROPOMI data a key step in identifying large, localized emission sources. Over more than seven years of operation, TROPOMI has performed a vast number of methane observations, which include many plume-like signals that cannot realistically be all identified through manual inspection. As a result, monitoring this growing volume of data for super-emitter activity requires automated detection methods, creating opportunities for machine learning approaches that depend on large and representative training datasets. However, detecting methane plumes in TROPOMI data remains challenging, as not every plume-like feature corresponds to an actual emission. Retrieval artifacts can closely resemble real methane plumes. Such artifacts may arise from factors including elevation or surface albedo gradients, high aerosol loading, coastlines, and the presence of water bodies. 

The Methane Hotspot Explorer is a new application launched by the Copernicus Atmospheric Monitoring Service (CAMS) \cite{copernicusCAMSMethane} to provide, on a weekly basis, large methane plumes detected worldwide by TROPOMI, thereby drawing global attention to the issue of large concentrated methane emissions. The application relies on the processing of TROPOMI methane data with a machine learning-based pipeline introduced by \cite{acp-23-9071-2023}. The first step of the pipeline is to identify methane plumes among all image patches (small spatial subsets) extracted from TROPOMI methane data. As a second step, another machine learning model is used to distinguish genuine methane plumes from retrieval artifacts. Finally, before being published on the Methane Hotspot Explorer, detections classified as plumes by the second machine learning model are double-checked manually by two independent expert labelers. Thus, the optimal performance of the plume-artifact classifier of this pipeline is crucial for reducing the human labor, as well as, maximizing the number of reported methane plumes.

In \cite{acp-23-9071-2023}, as the second step of the presented pipeline, the authors used a support vector machine (SVM) classifier trained on an extensive set of features designed by domain experts. The primary advantage of this approach is that multivariate modeling enables the extraction of relationships among variables that contribute to the formation of artifacts. However, this methodology also constrains the information available to the algorithm to the set of features deemed to be important by experts, disrupts the spatial relationships between pixels, and results in information loss through statistical aggregation.

In this study, we compare the performance of several feature-based and image-based classification models for the task of methane plume–artifact classification. We compare the performances of the models in two complementary regimes: balanced and imbalanced evaluation settings. In the imbalanced setting, we have more instances of one class in the dataset such as can be expected in a real-world application. The balanced setting, on the other hand, isolates the discriminative capacity of the models by removing class domination effects. In addition, we conduct a unified interpretability analysis for both feature-based and image-based models to understand the main factors contributing to the performance of both studied model types.

The paper is organized as follows: Section~\ref{litreview} reviews previous work on the topic of methane detection with machine learning. 
In Section~\ref{matmethods}, we explain our methodology by introducing the used data sources, machine learning concepts and the dataset. We present our results in Section~\ref{results} composed of metrics and the explainability analysis, and discuss the findings in Section~\ref{discussion}.

\section{Literature Review}
\label{litreview}
Various machine learning methods have been applied for tasks related to methane detection, ranging from feature-based models such as SVMs to deep neural networks applied to images \cite{acp-23-9071-2023, tiemann2025machine, chen2025mpsunet, kumar2020deep, bruno2024u}.
In \cite{acp-23-9071-2023}, the basis of this work, the authors applied an SVM to distinguish the emission plumes from retrieval artifacts on TROPOMI data. Other studies have also used data with higher spatial resolution, ranging from 0.5 to 60 meters, both satellite and airborne instruments. In \cite{chen2025mpsunet}, a UNet-based architecture model was used for segmenting methane plumes on data from the Earth Surface Mineral Dust Source Investigation (EMIT) instrument. A UNet model was also used in \cite{ruuvzivcka2023semantic} to segment methane plumes from a combination of hyperspectral and multispectral satellite images. In \cite{radman2023s2metnet}, the authors tested several deep learning architectures to enhance the precision of methane quantification using Sentinel-2 data. In \cite{kumar2020deep}, a Hyperspectral Mask-RCNN model was used to autonomously represent and detect methane plumes using data from the Airborne Visible/Infrared Imaging Spectrometer Next Generation (AVIRIS-NG). The proposed model outperformed feature-based models such as SVM and logistic regression. Finally, in \cite{zortea2023detection}, the authors showed the efficiency of a ResNet architecture for the task of classification between the Sentinel-2 images containing synthetic, simulated methane plumes and images that do not contain (synthetic) methane enhancements.

Our work extends beyond \cite{acp-23-9071-2023}, which is currently used for the CAMS Methane Hotspot Explorer. Specifically, for the task of plume-retrieval artifact classification on TROPOMI data, we compare deep learning image-based models (in particular, ResNet-18 and ResNet-34) with feature-based counterparts, including SVM as used in \cite{acp-23-9071-2023}, Random Forest, and Extreme Gradient Boosting (XGBoost). The latter two were included because they have shown superior performance over SVMs on similar TROPOMI-related tasks \cite{kurchaba2022supervised, kurchaba2023anomalous}. Thus, we aim to understand whether the application of more advanced machine learning techniques can potentially improve the workflow used in the Hotspot Explorer.

Furthermore, there are benefits in explaining machine learning models, especially when designing models to be deployed in production \cite{molnar2025}. Model explainability ensures that model outputs are reliable, or points out under which conditions a model is optimal for a given task \cite{molnar2025}. Therefore, in our study, we apply separate explainability techniques on feature-based and image-based models aiming at understanding the differences in learning processes of the used machine learning models. 

\section{Materials and Methods}
\label{matmethods}

\subsection{Machine learning methodology}

\subsubsection{Machine learning models}
The objective of this study is to compare the performance of feature-based classical machine learning models and image-based deep learning models for the classification of methane plumes and retrieval artifacts in TROPOMI satellite imagery. 
Feature-based models operate on hand-crafted or pre-extracted numerical descriptors derived from the TROPOMI images. These models do not process raw image data directly, but instead rely on feature representations designed to capture relevant spatial, spectral, or statistical characteristics of methane plume signatures. The following feature-based machine learning models are evaluated: Support Vector Classifier (SVC), which was selected to ensure direct comparability with the benchmark study \cite{acp-23-9071-2023}, Random Forest (RF) \cite{Breiman2001}, and  Extreme Gradient Boosting (XGBoost) \cite{chen2016xgboost}. All selected models are robust to noise and are able to model complex interactions between features.

In contrast to classical models, deep learning approaches are trained directly on TROPOMI image patches, allowing the models to directly capture spatial patterns and contextual information relevant to plume detection. In this study, we test two image-based architectures: ResNet-18 and ResNet-34 \cite{he2016deep}. Both models are variants of the residual neural network architecture, which incorporates skip connections to facilitate the training of deeper networks by mitigating the vanishing gradient problem \cite{vanishingcite}, where gradients become progressively smaller as they are backpropagated through many layers, making early layers difficult to train effectively. The use of two configurations of ResNet (18 and 34 layers) allows for an assessment of the impact of model capacity on classification performance. Initially designed for image recognition, the architectures are widely used in remote sensing \cite{zhao2022deep, wang2019scene}, and atmospheric measurements in particular. For instance, in \cite{hu2024high}, the authors used a ResNet architecture to estimate daily global carbon monoxide (CO) concentrations using TROPOMI data. In \cite{zhang2025enhancing}, a modification of the ResNet architectures was used to increase spatial resolution of NO$_2$ measurements from Ozone Monitoring Instrument (OMI), and in \cite{shi2024harmonizing}, ResNet was used as a benchmark for harmonizing ozone column concentration datasets from OMI and TROPOMI instruments.

To make sure that we exploit the maximum potential of a given machine learning model, we optimize the hyperparameters of each studied model. The hyperparameters are optimized using a random search technique \cite{randomsearch}, where the objective is to maximize average precision (defined in the following subsection) under 5-fold cross-validation. A set of hyperparameters yielding the best cross-validation result is applied on a hold-out test set. The list of hyperparameters optimized for each applied model, their search spaces, and the sets of hyperparameters yielding the best performance of the studied models can be found in Appendix \ref{app:A:hparams}.

\subsubsection{Model performance evaluation}

The performance of the classification models is evaluated using several complementary metrics.
First, the precision–recall (PR) curve is used. This curve depicts precision as a function of recall and is particularly informative for imbalanced datasets. Precision and recall are defined as:

\begin{equation}
\text{Precision} = \frac{TP}{TP + FP},
\end{equation}

\begin{equation}
\text{Recall} = \text{True Positive Rate} = \frac{TP}{TP + FN}.
\end{equation}

Here, $TP$ (true positives) corresponds to images containing a methane plume that were correctly identified by the classifier.
$FP$ (false positives) refers to images containing artifacts that were incorrectly classified as containing a methane plume.
$FN$ (false negatives) denotes images depicting a methane plume that were incorrectly classified as artifacts.

To summarize the information contained in the PR curve into a single scalar value, average precision (AP) is computed. Average precision corresponds to the area under the precision–recall curve and reflects the trade-off between precision and recall across different decision thresholds. 

In addition to PR-based metrics, Receiver Operating Characteristic (ROC) analysis is performed. The ROC curve visualizes the true positive rate as a function of the false positive rate, defined as:

\begin{equation}
\text{True Positive Rate} = \text{Recall} = \frac{TP}{TP + FN},
\end{equation}

\begin{equation}
\text{False Positive Rate} = \frac{FP}{FP + TN},
\end{equation}

where $TN$ denotes true negatives, corresponding to artifact images correctly classified as such. The area under the ROC curve (ROC-AUC) is used as a threshold-independent measure of the model’s discriminative ability.

Finally, balanced accuracy is used as an additional evaluation metric to evaluate model performance a specific operating point. Balanced accuracy is defined as the average of sensitivity (true positive rate) and specificity (true negative rate):

\begin{equation}
\text{Balanced Accuracy} = \frac{1}{2}\left(\frac{TP}{TP + FN} + \frac{TN}{TN + FP}\right).
\end{equation}

The used balanced version of the accuracy metric provides a more representative estimate of performance when the number of samples in each class is unequal, as it gives equal weight to both classes.

\subsection{Data}
\subsubsection{Data sources}
The primary data source used in this study is the TROPOspheric Monitoring Instrument (TROPOMI) \cite{veefkind2012tropomi} on board ESA's Sentinel-5P satellite. TROPOMI provides daily global observations of atmospheric methane with pixel sizes down to approximately 7 km $\times$ 5.5 km at nadir and a point-source detection limit of about 8 tons h$^{-1}$ under favorable conditions \cite{acp-23-9071-2023}. We use the operational TROPOMI Level-2 methane product version up to 02.06.00 distributed through the Copernicus Data Space \cite{tropomilevel2methane}.

From this product and associated auxiliary inputs, we use both methane-retrieval variables and geophysical/context variables. The methane-related variables include methane mixing ratio XCH$_4$, bias-corrected methane mixing ratio, and methane mixing ratio precision. Retrieval diagnostics variables include QA value and $\chi^2$ (fit residual). Context variables include SWIR surface albedo, SWIR aerosol optical thickness, surface pressure, surface altitude, cloud fraction (VIIRS SWIR IFOV), surface classification, and near-surface wind components (eastward and northward; ECMWF 10U/10V) \cite{apituley2021sentinel}. 

\subsubsection{Dataset}
We focus on the binary classification task of distinguishing TROPOMI image patches containing methane plumes from patches containing retrieval artifacts.
To obtain an initial set of detections, we follow the plume detection pipeline described by \cite{acp-23-9071-2023}, which represents the current state-of-the-art for continuous automatic detection of methane plumes from large super-emitters in TROPOMI data. This pipeline also forms the basis of the CAMS Methane Hotspot Explorer \cite{copernicusCAMSMethane}.

The first stage of the pipeline uses a classification model from \cite{acp-23-9071-2023} that identifies plume-like objects in all TROPOMI methane observations, which are split into 32x32 image patches. 
We use this model to generate the initial set of detections. 
The second step of \cite{acp-23-9071-2023} aggregates TROPOMI measurements and supplementary data into one-dimensional features, which are then used to train a classification model capable of distinguishing methane plumes from retrieval artifacts in the detections of the first model. 
We adopt the feature set proposed by \cite{acp-23-9071-2023} for the feature-based models (SVC, RF, XGBoost). 
A full list of these features, along with brief descriptions, is provided in Appendix \ref{A: features}. 
For detailed information on feature derivation, we refer the reader to the original paper.

For the image-based models (ResNet-18 and ResNet-34), we use methane concentration image patches, as in the first stage of the pipeline, together with supplementary channels such as surface albedo, surface altitude, aerosol optical thickness (AOT), surface pressure, snow cover (blended surface albedo) \cite{lorente2021methane}, wind, and surface classification. We also include a plume mask channel. This channel encodes which pixels of the patch compose a plume as detected by the first model. We multiply the binary mask with the score of the first CNN model (probability of an image patch containing a plume). Therefore, if the plume mask is encoded with values close to 1, it indicates that the given plume was detected by the first model with high confidence. However, if the plume mask is encoded with values near 0, it indicates a low confidence of detection. The complete list of features used to train the image-based models, along with their descriptions, can be found in Appendix \ref{A: features_ResNet}.

\begin{table}
  \centering
    \begin{tabular}{cccc}
    \hline
      \textbf{Count} & \textbf{Training} & \textbf{Test}\\
     \hline
    \textbf{Plumes} & 6085 & 276 \\
    \textbf{Artifacts} & 2798 & 286\\
    \textbf{Total} & 8883 & 562\\ 
    \hline
    \end{tabular}
    \caption{Number of data points used in training and test sets.}
\label{tab:dataset_count}
\end{table}

\begin{figure*}
    \centering
    \includegraphics[width=0.9\linewidth]{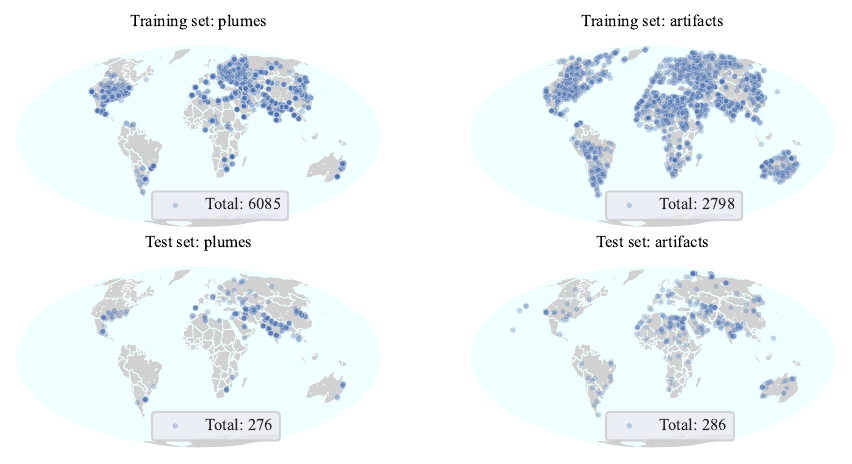}
    \caption{Spatial distribution of data points from the training and test sets.}
    \label{fig:spatial_distrib}
\end{figure*}

Because image-based models require significantly larger datasets than feature-based models, using the dataset presented in \cite{acp-23-9071-2023} as a basis, we created expanded datasets both for feature-based and image-based models. 
While the original training set for feature-based models contained 843 image patches, our dataset comprises 8,883 patches, including 6,085 methane plumes and 2,798 retrieval artifacts. This expansion was achieved by adding real image patches of newly detected plumes and artifacts, rather than by applying artificial data augmentation techniques. 
Table \ref{tab:dataset_count} provides further details on the dataset. 

As shown in Table \ref{tab:dataset_count}, the training set exhibits a moderate class imbalance, which reflects real-world detection scenarios where the ratio between two classes can vary. 
This experiment design allows us to evaluate the intrinsic discriminative ability of the models while minimizing the influence of class dominance, ensuring a fair comparison across different model architectures.
The dataset spans the period from April 2019 to August 2024 and has global spatial coverage. The spatial distribution of the training and test sets is illustrated in Figure \ref{fig:spatial_distrib}.

\section{Results}
\label{results}

\begin{table} 
    \centering 
    \begin{tabular}{cccc} 
    \hline
    \textbf{Model} & \textbf{Average Precision} & \textbf{ROC-AUC} & \makecell{\textbf{Balanced}\\\textbf{Accuracy}} \\ 
    \hline
     SVC & 0.918 $\pm$ 0.012 & 0.869 $\pm$ 0.017 & 0.768 $\pm$ 0.024 \\ 
     RF  & 0.948 $\pm$ 0.008 & 0.908 $\pm$ 0.014 & 0.809 $\pm$ 0.015 \\ 
     XGB & 0.944 $\pm$ 0.013 & 0.905 $\pm$ 0.018 & 0.801 $\pm$ 0.022 \\ 
     ResNet-18 & 0.905 $\pm$ 0.047 & 0.869 $\pm$ 0.044 & 0.771 $\pm$ 0.066 \\ 
     ResNet-34 & 0.918 $\pm$ 0.034 & 0.888 $\pm$ 0.043 & 0.829 $\pm$ 0.055 \\ 
     \hline
     \end{tabular} 
     \caption{Aggregated mean $\pm$ standard deviation from 5-fold cross-validation with the best set of hyperparameters.} 
    \label{tab:cv_results}
\end{table}

\begin{table}
    \centering
    \begin{tabular}{cccc}
    \hline
      \textbf{Model} & \textbf{Average Precision} & \textbf{ROC-AUC} & \makecell{\textbf{Balanced}\\\textbf{Accuracy}} \\
     \hline
        SVC           & 0.919 & 0.928 & 0.843 \\
        RF            & 0.917 & 0.927 & 0.852 \\
        XGB           & 0.922  & 0.929 & 0.859 \\
        ResNet-18     & 0.937 & 0.939 & 0.847 \\
        ResNet-34     & 0.929 & 0.934 & 0.836\\
    \hline
    \end{tabular}
     \caption{Results on the hold-out test set.}
    \label{tab:test_results}
\end{table}

\begin{figure*}
    \centering
    \includegraphics[width=1.0\linewidth]{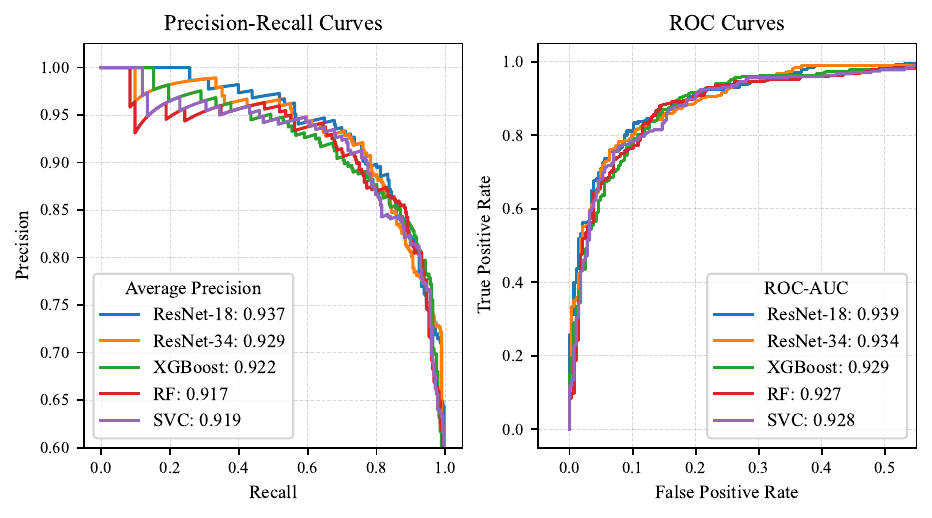}
    \caption{Precision-recall (left panel) and ROC-AUC (right panel) curves on the hold-out test set for the studied models.}
    \label{fig:prec_recall}
\end{figure*}

We analyze the results under two complementary evaluation settings: an imbalanced setting that reflects real-world conditions and a balanced setting that isolates the intrinsic discriminative ability of the models.
First, we consider the imbalanced setting, corresponding to the 5-fold cross-validation results reported in Table \ref{tab:cv_results}, where the validation folds preserve the original class ratio of approximately 1:2. In this scenario, classical machine learning models, and in particular tree-based ensembles, exhibit the strongest and most stable performance. The RF model achieves the highest average precision (0.948 ± 0.008; 5-fold average and standard deviation) and the best ROC-AUC (0.908 ± 0.014), while also maintaining a competitive balanced accuracy. XGBoost shows similarly strong performance across all metrics, with means that are slightly lower but within the observed fold-to-fold variability. These results indicate that ensemble methods are particularly robust under class imbalance and limited effective training data per fold. Meanwhile, the SVC model used in \cite{acp-23-9071-2023} (but trained on the same data) falls behind the other feature-based models, which is expectable, as it is typically less powerful.  The image-based models (ResNet-18 and ResNet-34) achieve lower average precision and ROC-AUC in cross-validation. However, their larger standard deviations suggest that there may be settings where they can be competitive against the feature-based models.

One such case is the balanced evaluation setting on the hold-out test set. In this experiment, each model was evaluated using the hyperparameter configuration selected during the preceding cross-validation phase. The results are summarized in Table \ref{tab:test_results} and further illustrated by the precision–recall and ROC curves in Figure \ref{fig:prec_recall}. In this setting, the relative ranking of models changes noticeably. Based on Table \ref{tab:test_results}, ResNet-18 achieves the highest average precision (0.937) and the highest ROC-AUC (0.939), slightly outperforming all feature-based approaches. Although XGBoost achieves the highest balanced accuracy on the test set, its average precision and ROC-AUC remain lower than those of ResNet-18/34, indicating that its advantage is limited to a specific operating point rather than its overall discriminative performance. Indeed, this can be seen by analyzing the precision–recall curves in Figure \ref{fig:prec_recall}, where ResNet-18 consistently maintains higher precision across most recall values, particularly in the high-recall regime. ResNet-34 also performs strongly but does not surpass ResNet-18, suggesting that additional depth does not yield further gains for this task.

Overall, all models achieve strong performance, indicating that the task of plume-artifact classification is learnable across a range of model families. Depending on the training and evaluation settings, some models perform better than others. In particular, on the imbalanced setting, feature-based models showed strong performances with little variability, while the image-based ones performed better on the balanced hold-out test set. The next Subsection discusses the type of insights we can extract out of the models.

\subsection{Model explainability}
In this Subsection, we perform an explainability analysis of the presented models. We employ SHapley Additive exPlanations (SHAP) as it is a unified framework for explainability across different model families, and is therefore applicable to our comparative study. SHAP is grounded in cooperative game theory and attributes a prediction to individual input features by estimating their marginal contributions to the model output.

\subsubsection{Tree-based model}
\begin{figure*}
    \centering
    \includegraphics[width=0.9\linewidth]{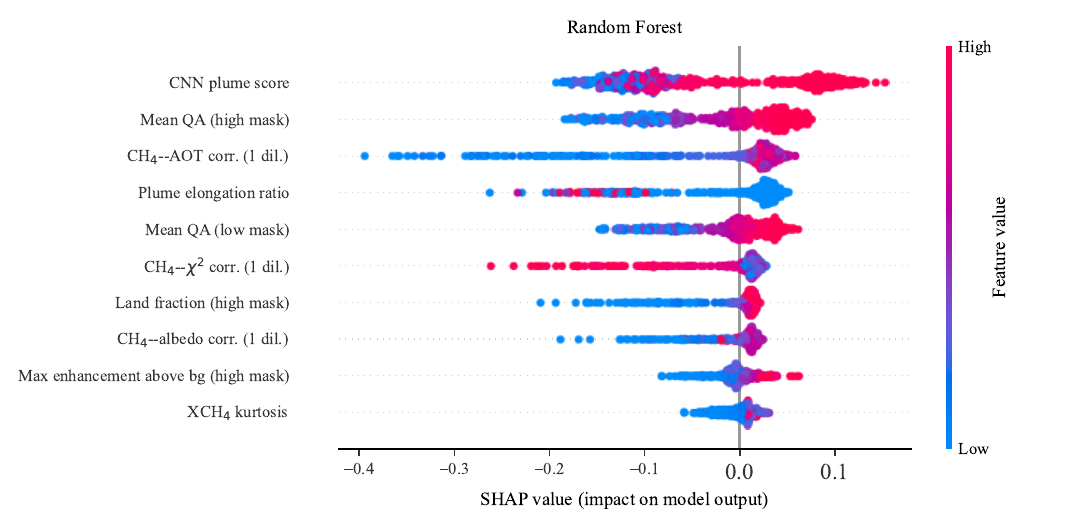}
    \caption{SHAP summary plot for the Random Forest model. The ten most important variables are shown, where importance is defined by the mean absolute SHAP value across all samples. AOT stands for aerosol optical thickness. Bg stands for background.}
    \label{fig:shap_plot}
\end{figure*}

For the tree-based models (RF, XGboost), we use the TreeExplainer \cite{Lundberg2020}, which enables estimation of SHAP values by exploiting the structure of decision trees. Feature relevance is summarized using SHAP summary plots, which aggregate local explanations over the dataset and display both the magnitude and direction of feature contributions. In this context, the importance of a feature is measured as the mean absolute SHAP value across all samples, i.e., the average absolute change in the model output attributable to that feature relative to the model baseline.
Figure \ref{fig:shap_plot} presents a SHAP summary plot for the RF model. The features are ordered from top to bottom by decreasing overall contribution to the model prediction, as quantified by this mean absolute SHAP value. The ten most important features are depicted in the figure.
Each point represents an individual sample, with its horizontal position corresponding to the SHAP value. Positive SHAP values indicate that the predicted probability is pushed toward the positive class, whereas negative values indicate the contrary. Point color denotes the feature value, ranging from low (blue) to high (red).

The most influential features are \textit{CNN plume score}, \textit{Mean QA (high mask)}, and \textit{CH$_4$--AOT corr. (1 dil.)}, with the \textit{CNN plume score} being on top of the list. This feature tells how confident the first (plume - not a plume) classification model was in classifying the image as containing a plume. Higher values of \textit{CNN plume score} are predominantly associated with positive SHAP contributions, indicating that stronger CNN-derived confidence consistently drives the model toward the positive class (methane plume).
In contrast, for the feature \textit{CH$_4$--AOT corr. (1 dil.)}, we see a strong influence of the low values of the feature on the negative response of the RF model. This feature corresponds to the Pearson correlation between the CH$_4$ concentration and aerosol optical thickness for pixels within one dilation around the low confidence plume mask, where a high value typically indicates the detection is an artifact. Mid-ranked features, including \textit{Plume elongation ratio}, \textit{Mean QA (low mask)}, \textit{CH$_4$--$\chi^2$ corr. (1 dil.)}, \textit{Land fraction (high mask)}, \textit{CH$_4$--albedo corr. (1 dil.)}, show moderate but consistent contributions to the model responses, reflecting the reliance of the model on complementary features related to wind, quality of the retrievals, albedo, or surface conditions. Features like \textit{Max enhancement above bg (high mask)} and \textit{XCH$_4$ kurtosis} exhibit comparably small SHAP magnitudes, indicating a limited influence on the final predictions.

\begin{table*}
    \centering
    \begin{tabular}{ccc}
    \hline
             & \textbf{RF} & \textbf{XGBoost} \\
     \hline
        1 & CNN plume score & CNN plume score \\
        2 & Mean QA (high mask) & CH$_4$--AOT corr. (1 dil.) \\
        3 & CH$_4$--AOT corr. (1 dil.) & Mean QA (high mask) \\
        4 & Plume elongation ratio & Plume elongation ratio \\
        5 & Mean QA (low mask) & CH$_4$--$\chi^2$ corr. (1 dil.) \\
        6 & CH$_4$--$\chi^2$ corr. (1 dil.) & CH$_4$--albedo corr. (1 dil.) \\
        7 & Land fraction (high mask) & Land fraction (high mask) \\
        8 & CH$_4$--albedo corr. (1 dil.) & Mean QA (low mask) \\
        9 & Max enhancement above bg (high mask) & 10 m wind speed\\
        10& XCH$_4$ kurtosis & Plume--wind angle\\
    \hline
    \end{tabular}
    \caption{The ten most important variables for RF and XGBoost models ranked from highest to lowest.}
    \label{tab:var_importance}
\end{table*}

In Table \ref{tab:var_importance}, we list the ten most important features for RF and XGBoost models identified using the SHAP TreeExplainer presented earlier. The most relevant features of both models overlap and have similar rankings, which is consistent with the comparable performance of both models across balanced and imbalanced evaluation settings. At the same time, some noteworthy differences emerge: the XGBoost list includes such variables as \textit{10 m wind speed} and \textit{Plume--wind angle}, whereas RF instead includes \textit{Max enhancement above bg (high mask)} and \textit{XCH$_4$ kurtosis}. These differences suggest that XGBoost relies relatively more on wind-related context, while RF gives relatively more weight to enhancement-shape statistics. These differences may be partly explained by the structural properties of the models. The RF model relies on feature subsampling and tends to select among correlated predictors somewhat interchangeably, often favoring features with strong standalone discriminative power, such as enhancement statistics. In contrast, XGBoost builds trees sequentially and can exploit correlated or weakly predictive variables more effectively when they provide incremental improvements, particularly through interactions. This may explain the stronger reliance of XGBoost on wind-related context variables, which likely contribute in combination with plume geometry rather than as independent predictors. While this interpretation is consistent with known differences between bagging and boosting methods used in RF and XGBoost respectively, further analysis is required to confirm this behavior.

\subsubsection{Image-based model}
Building on the SHAP analysis of the tree-based model, we extend the explainability study to the image-based deep learning models, focusing on a ResNet-18 architecture. While SHAP values in tree-based models directly quantify feature contributions at the input level, explaining deep neural networks requires accounting for hierarchical feature extraction (across multiple convolutions) and spatial dependencies. 
We employ an integrated SHAP GradientExplainer \cite{Lundberg2017} to analyze both channel-level importance and pixel-level attributions. 
SHAP GradientExplainer approximates SHAP values by computing input gradients and integrating them along a continuous path from a reference input to the actual sample. In our case, a reference input is composed of a subset of 50 randomly selected images, which is a trade-off between the stability of the obtained results and computation speed.
These integrated attributions are aggregated in two complementary ways: globally across channels (mean absolute SHAP values) and locally in image space (regions that push predictions toward plume or artifact classes). This approach is well-suited for deep neural networks, as it accounts for nonlinearities and interactions inherent in convolutional architectures while remaining computationally tractable for high-dimensional, image-based inputs.

\begin{figure*}
    \centering
    \includegraphics[width=.8\linewidth]{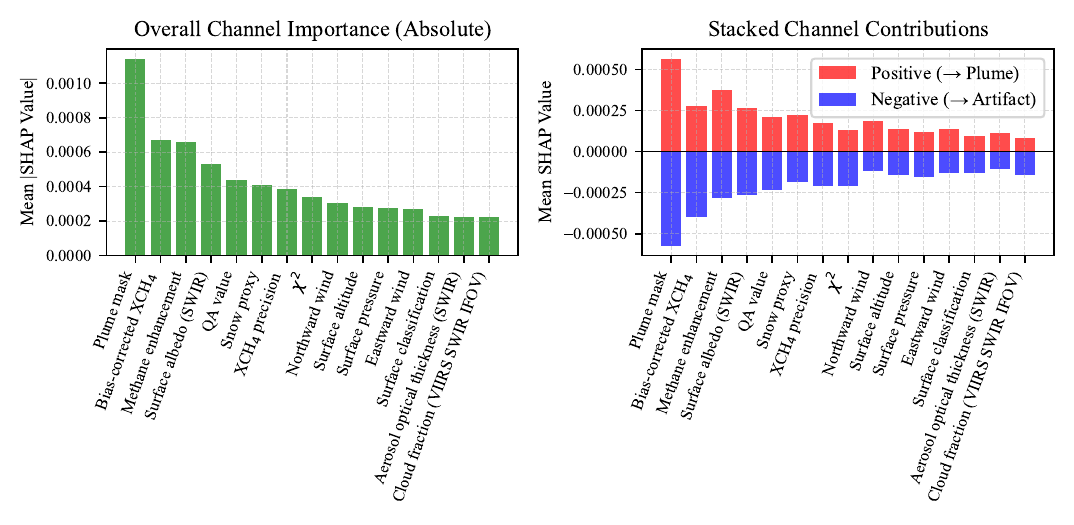}
    \caption{Integrated SHAP Gradient Explainer for ResNet-18. Left panel: mean absolute SHAP values per channel. Right panel: mean positive and negative SHAP values per channel.}
    \label{fig:images_shap}
\end{figure*}

Figure \ref{fig:images_shap} summarizes the global importance of the input channels for the ResNet-18 model. 
The left panel reports the mean absolute SHAP value per channel, providing a measure of overall contribution irrespective of sign. 
The most important channel for the ResNet-18 model is the \textit{plume mask} channel. 
As described earlier, the plume mask channel encodes the \textit{CNN plume score}, which was identified as the most important variable in the tree-based model.
Moreover, the plume mask channel delineates the spatial structure of the detected plume-like object, which is particularly important in the context of image-based learning. 
In contrast, channels related to QA values or aerosol concentration, derivatives of which were highly relevant for the tree-based models, appear in the middle or toward the end of the ranking for the ResNet-18 model, indicating a more secondary or contextual role. While surface albedo, which did not have a strong influence on the tree-based models, appears as one of the most important features of ResNet-18.

The right panel of Figure~\ref{fig:images_shap} decomposes channel contributions into mean positive and mean negative SHAP values, highlighting how individual channels support or oppose a given class prediction. In contrast to the tree-based model, where feature effects were mostly unidirectional, the ResNet-18 model exhibits more balanced positive and negative contributions within the same channel. This behavior reflects the spatially localized and context-dependent nature of convolutional features, where the same input channel may provide evidence for different classes depending on spatial patterns and interactions with other channels (in our case, methane enhancement in particular).

Comparing the explainability results of image-based and feature-based models reveals an important difference in how albedo information is used. In the image-based ResNet model, \textit{surface albedo} ranks among the most important input channels, suggesting that the model may have learned that spatial correspondence between albedo patterns and methane enhancements is informative for identifying retrieval artifacts. In contrast, in the tree-based models, albedo appears only as a secondary contextual variable, which may suggest that information contained in this feature is not utilized to the full extent. 

\begin{figure*}[t]
     \centering
    \includegraphics[width=1.0\linewidth]{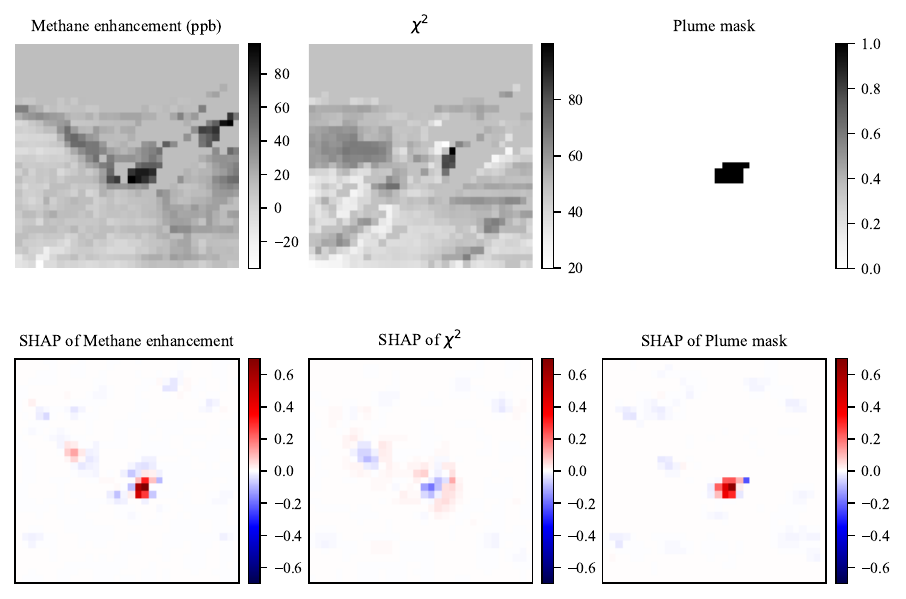}
    \caption{Image classified as "plume". Top row: values of the selected model channels. Bottom row: local pixel-wise explanations of the corresponding channel.}
    \label{fig:shap_image_plume}
\end{figure*}

\begin{figure*}[t]
     \centering
    \includegraphics[width=1.0\linewidth]{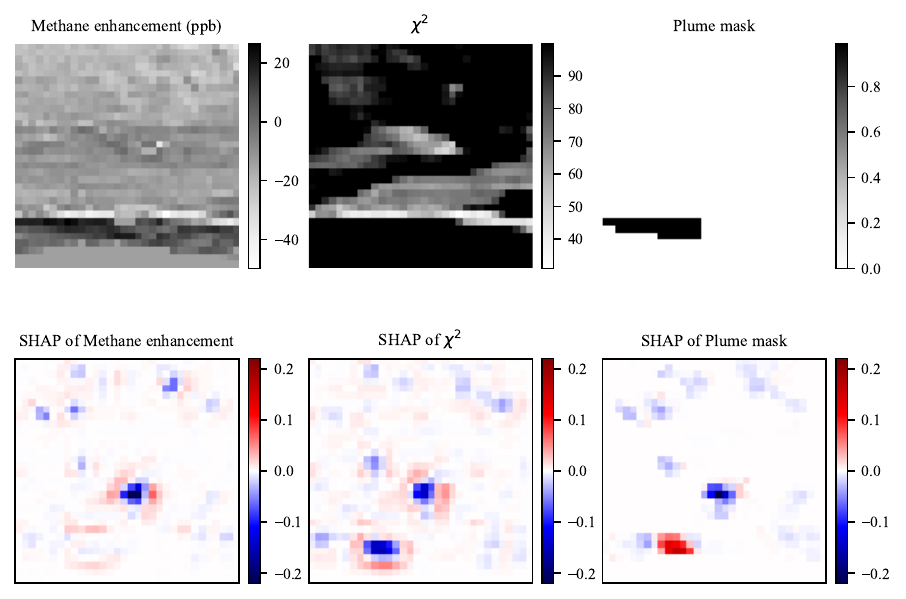}    
    \caption{Image classified as "artifact". Top row: values of the selected model channels. Bottom row: local pixel-wise explanations of the corresponding channel.}
    \label{fig:shap_image_artifact}
\end{figure*}

To complement the global channel-level analysis, Figures~\ref{fig:shap_image_plume} and~\ref{fig:shap_image_artifact} provide local, pixel-wise explanations for representative examples. Figure~\ref{fig:shap_image_plume} shows an image classified as a "plume". The top row corresponds to the selected input channels, while the bottom row visualizes positive and negative SHAP values for individual pixels of the corresponding channel. The attribution maps show that the model assigns strong positive contributions to spatially coherent regions corresponding to the plume structure in channels such as \textit{methane enhancement} and \textit{plume mask}. 

Figure~\ref{fig:shap_image_artifact} presents an example classified as an "artifact". In this case, the $\chi^{2}$ channel contributes negatively in the region where a plume-like signal is expected based on the \textit{plume mask} channel. Moreover, in contrast to Figure~\ref{fig:shap_image_plume}, the magnitude of gradients is much lower and more balanced between positive and negative contributions. 

Overall, the SHAP analysis of the ResNet-18 model reveals a strong conceptual alignment with the findings from the tree-based model, while also highlighting key differences that are the consequence of spatial learning. 
Both models identify plume-related information as the main driver of classification decisions. However, image-based deep learning models naturally exploit spatial coherence and localized patterns, while tree-based model compensate missing spatial information by relying on more additional features.  

\section{Discussion}
\label{discussion}

This study systematically evaluated classical feature-based and image-based models under two complementary settings: balanced and imbalanced class distribution.
The imbalanced setting revealed a clear advantage for feature-based approaches. In particular, the Random Forest (RF) model demonstrated consistently strong and stable performance, showcasing its robustness in such settings.

In contrast, the balanced evaluation setting isolated the discriminative capacity of the models by removing class domination effects. Under these conditions, the ResNet-18 architecture outperformed other approaches, achieving the highest average precision and ROC-AUC. This result highlights the ability of deep convolutional models to utilize spatial representations when sufficient and well-balanced training data is available. Importantly, the highest ROC-AUC and average precision of ResNet-18 suggest better flexibility in threshold selection, which is critical in settings where the proper selection of the trade-off between false positives and false negatives is important. Overall, these findings show the complementary strengths of both model types: RF/XGB is preferable when robustness under imbalance is the main requirement, while ResNet-18 is optimal when the main goal is to maximize the discriminative performance of the model.

In addition to predictive performance, model interpretability is important for building trust in the results coming from machine learning, as well as gaining scientific insight from the obtained results. We performed a unified explainability analysis using SHapley Additive exPlanations (SHAP), enabling direct comparison across feature-based and image-based models. For feature-based models, SHAP analysis identified the \textit{CNN plume score} as the most influential feature. The dominance of the \textit{CNN plume score} indicates that the confidence of the first-stage plume-detection CNN plays a central role in second-stage "plume-artifact" classification. Other physical or quality-related variables show moderate but consistent contributions to model responses.

The explainability analysis of the ResNet-18 model further confirms these conclusions while also showing differences specific to spatial learning. Using an integrated SHAP GradientExplainer, we studied both channel-level importance and pixel-level attributions. As with the tree-based models, the \textit{CNN plume score} (integrated into the plume mask channel) turned out to be the most influential input, confirming that the confidence of the first-stage plume-detection is the main driver of classification decisions for both model families. However, unlike tree-based models, which depend more heavily on auxiliary scalar features to compensate for the absence of spatial context, ResNet-18 takes advantage of the spatial structure encoded in the plume mask and the related methane concentration. This allows the model to capture the shape, coherence, and localization clues that are essential for image-based discrimination.

From an application perspective, we interpret the obtained results in the context of the Methane Hotspot Explorer pipeline, whose primary objective is the reliable and timely identification of methane plumes while minimizing the burden on human experts, who verify all positive detections in a quality control process. In this operational setting, the second-stage plume–artifact classifier plays a crucial role: any false positive directly increases the manual verification workload, whereas false negatives lead to missed emission events and reduced scientific and societal impact. The observed robustness of tree-based models under a class imbalance setting and limited effective training data becomes a crucial characteristic of the model needed for effective functioning of the Methane Hotspot Explorer pipeline. 

At the same time, ResNet-18 showed a slightly higher discriminative performance in the balanced evaluation setting. This suggests the potential of image-based deep learning to further advance the pipeline when the main goal is maximizing plume detection capability. The high ROC-AUC achieved by the ResNet-18 shows better flexibility in threshold tuning, which could be leveraged to adapt the system to evolving operational priorities—for example, prioritizing sensitivity during targeted emission monitoring campaigns or precision when expert resources are limited. Moreover, this gain is achieved by directly exploiting spatial information in the TROPOMI-derived inputs, rather than relying exclusively on handcrafted statistical features. Finally, the image-based structure of the model allows for an easy integration of the second-stage plume-artifact model with the first-stage CNN used for an initial detection of plume-like objects.

Summing up, we found settings where tree-based and image-based machine learning models demonstrate competitive performances against each other, and related them to operational scenarios.
 Further study can build upon our work by increasing the complexity of the approaches, whether it is by designing more features or applying other deep learning architectures for plume-artifact discrimination. The importance of the first-stage pipeline model suggests that further improvement of the Methane Hostspot Explorer methodology could be focused on this part of the pipeline. Finally, the presented explainability methods can be used for finding insights regarding the distinguishability between methane plumes and artifacts that are not yet known to experts.

\begin{acks}
This work was produced with funding by the Copernicus Atmosphere Monitoring Service implemented by the European Centre for Medium-Range Weather Forecasts (ECMWF) on behalf of the European Commission.
\end{acks}

\typeout{}

\appendix
\begin{appendices}

\section{Features used in feature-based models}\label{A: features}
\begin{itemize}
    \item \textit{CNN plume score} - Prediction score assigned by the CNN from the first step of the pipeline.
    \item \textit{Valid pixel fraction} - Fraction of valid pixels in the scene, N/(32×32).
    \item \textit{High-mask pixel count} - Number of pixels in the high-confidence plume mask.
    \item \textit{High-mask enhancement sum} - Sum of enhancement of the pixels in the high-confidence plume mask.
    \item \textit{XCH$_4$ std. dev.} - Standard deviation of the XCH4 value of all pixels in the scene.
    \item \textit{XCH$_4$ skewness} - Skewness of the XCH4 value of all pixels in the scene.
    \item \textit{XCH$_4$ kurtosis} - Kurtosis of the XCH4 value of all pixels in the scene.
    \item \textit{IME [kg]} - Integrated mass enhancement.
    \item \textit{Plume length} - Plume length, computed as sqrt(Area), following the IME equation definition \cite{varon2018quantifying}.
    \item \textit{10 m wind speed} - Wind speed, obtained from ERA5 \cite{hersbach2020era5}, present in the L2 methane product.
    \item \textit{Cloud-adjacent enhancement sum} - Sum product of a 3×3 kernel multiplying enhancements of the high-confidence plume mask with the cloud fraction.
    \item \textit{Cloud-adjacent pixel count} - Number of pixels in the high-confidence plume mask close to cloudy pixels based on a 3×3 kernel.
    \item \textit{Plume--wind angle} - Angle between the principal axis of the plume mask and the mean wind vector.
    \item \textit{Plume elongation ratio} - Ratio between the variance along the primary and secondary axis.
    \item \textit{CH$_4$--albedo corr. (scene)} - Pearson r value between XCH4 and albedo for the full scene.
    \item \textit{CH$_4$--albedo corr. (1 dil.)} - Pearson r value between XCH4 and albedo for pixels within one dilation around the low-confidence mask.
    \item \textit{CH$_4$--AOT corr. (scene)} - Pearson r value between XCH4 and aerosol optical thickness (AOT) for the full scene.
    \item \textit{CH$_4$--AOT corr. (1 dil.)} - Pearson r value between XCH4 and AOT for pixels within one dilation around the low-confidence mask.
    \item \textit{CH$_4$--surface pressure corr. (scene)} - Pearson r value between XCH4 and surface pressure for the full scene.
    \item \textit{CH$_4$--surface pressure corr. (1 dil.)} - Pearson r value between XCH4 and surface pressure for pixels within one dilation around the low-confidence mask.
    \item \textit{CH$_4$--$\chi^2$ corr. (scene)} - Pearson r value between XCH4 and $\chi^2$ for the full scene.
    \item \textit{CH$_4$--$\chi^2$ corr. (1 dil.)} - Pearson r value between XCH4 and $\chi^2$ for pixels within one dilation around the low-confidence mask.
    \item \textit{Cloud angle (high mask)} - Angle of the principal axis of the high-confidence plume mask with the principal axis of a cloud.
    \item \textit{Cloud angle (low mask)} - Angle of the principal axis of the low-confidence plume mask with the principal axis of a cloud.
    \item \textit{Coast angle} - Angle of the principal axis of the high-confidence plume mask with a coast.
    \item \textit{Mean $\chi^2$ (high mask)} - Average $\chi^2$ value of the pixels within the high-confidence plume mask.
    \item \textit{Mean $\chi^2$ (low mask)} - Average $\chi^2$ value of the pixels within the low-confidence plume mask.
    \item \textit{Mean albedo (high mask)} - Average albedo value of the pixels within the high-confidence plume mask.
    \item \textit{Mean albedo (low mask)} - Average albedo value of the pixels within the low-confidence plume mask.
    \item \textit{Mean AOT (high mask)} - Average AOT value of the pixels within the high-confidence plume mask.
    \item \textit{Mean AOT (low mask)} - Average AOT value of the pixels within the low-confidence plume mask.
    \item \textit{Mean QA (high mask)} - Average QA value of the pixels within the high-confidence plume mask.
    \item \textit{Mean QA (low mask)} - Average QA value of the pixels within the low-confidence plume mask.
    \item \textit{Background XCH$_4$ std. dev. (high mask)} - Standard deviation of the XCH4 values of pixels outside of the high-confidence plume mask, similar to the pixel precision of \cite{varon2021high}.
    \item \textit{Background XCH$_4$ std. dev. (low mask)} - Standard deviation of the XCH4 values of pixels outside of the low-confidence plume mask, similar to the pixel precision of \cite{varon2021high}.
    \item \textit{Mean enhancement above bg (high mask)} - Average enhancement above the background of the pixels within the high-confidence plume mask.
    \item \textit{Mean enhancement above bg (low mask)} - Average enhancement above the background of the pixels within the low-confidence plume mask.
    \item \textit{Max enhancement above bg (high mask)} - Maximum enhancement above the background of the pixels within the high-confidence plume mask.
    \item \textit{Land fraction (high mask)} - Fraction of pixels with surface classification “land” in the high-confidence plume mask.
    \item \textit{Land+water fraction (high mask)} - Fraction of pixels with surface classification “land+water” in the high-confidence plume mask.
    \item \textit{Coast fraction (high mask)} - Fraction of pixels with surface classification “coast” in the high-confidence plume mask.
\end{itemize}

\section{Features used in image-based models}\label{A: features_ResNet}
\begin{itemize}
    \item \textit{Bias-corrected XCH$_4$} - Bias- and stripe corrected column-averaged dry-air mole fraction of methane \cite{amt-11-5507-2018}.
    \item \textit{Methane enhancement} - Methane enhancement relative to the local background \cite{acp-23-9071-2023}, used to highlight plume-like signals.
    \item \textit{XCH$_4$ precision} - Retrieval precision of the XCH$_4$ estimate at pixel level.
    \item \textit{Surface albedo (SWIR)} - Shortwave-infrared surface reflectance from by the Level-2 retrieval, related to surface brightness conditions.
    \item \textit{Aerosol optical thickness (SWIR)} - Retrieved aerosol optical thickness in the SWIR band, affecting light path and retrieval quality.
    \item \textit{$\chi^2$} - Retrieval spectral fit residual metric; higher values indicate a poorer fit between the forward model and the measurements.
    \item \textit{Surface altitude} - The mean of the sub-pixels of the surface altitude within the field of view that is defined by the pixel corner coordinates. The source data is the Copernicus 90 m surface elevation database \cite{apituley2021sentinel}.
    \item \textit{Surface pressure} - Pressure at the surface elevation of the S5P SWIR pixel \cite{apituley2021sentinel}.
    \item \textit{QA value} - Quality descriptor, varying between 0 (no data) and 1 (best quality data). The value will change based on observation conditions and retrieval flags \cite{apituley2021sentinel}.
    \item \textit{Eastward wind} - The horizontal component of the wind at 10-meter height in the eastward direction \cite{hersbach2020era5}. This is the 10U parameter from ECMWF \cite{apituley2021sentinel}.
    \item \textit{Northward wind} -  The horizontal component of the wind at 10-meter height in the northward direction \cite{hersbach2020era5}. This is the 10V parameter from ECMWF \cite{apituley2021sentinel}.
    \item \textit{Snow proxy} - Indicator of snow/ice-covered surfaces, based on the blended surface albedo \cite{lorente2021methane}.
    \item \textit{Surface classification} - Discrete land-surface type class (e.g., land, coast, water) associated with each pixel.
    \item \textit{Plume mask} - A binary plume mask (plume pixels/non-plume pixels) multiplied with the score of the first CNN model (probability
of an image patch containing a plume).
    \item \textit{Cloud fraction (VIIRS SWIR IFOV)} - Cloud fraction from VIIRS data in the SWIR channel for the instantaneous field of view \cite{apituley2021sentinel}. 
\end{itemize}

\section{Hyperparameters' search space and optimal hyperparameters}
\label{app:A:hparams}

\begin{table*}[t]
\centering
\begin{tabular}{llp{7cm}l}
\hline
\textbf{Hyperparameter} & \textbf{Search Space} & \textbf{Description} & \textbf{Best found} \\
\hline
$C$ & $\mathrm{LogUniform}(10^{-3}, 10^{3})$ & 
Regularization parameter controlling the trade-off between margin maximization and classification error. & $10.564$ \\

Kernel & \{rbf, linear, poly\} & 
Specifies the kernel type used in the SVM algorithm. & rbf \\

$\gamma$ & $\mathrm{LogUniform}(10^{-4}, 1)$ & 
Kernel coefficient for `rbf` and `poly` kernels. Determines the influence of individual training samples. & $0.0027$ \\

Degree & \{2, 3, 4\} & 
Degree of the polynomial kernel function (used only when kernel = poly). & $4$ \\
\hline
\end{tabular}
\caption{Random search hyperparameter space and best-found hyperparameters for the Support Vector Classifier (SVC). We used the SVC model implementation from scikit-learn v.1.8.0.}
\label{tab:svc_random_search}
\end{table*}

\begin{table*}[h]
\centering
\begin{tabular}{llp{7cm}l}
\hline
\textbf{Hyperparameter} & \textbf{Search Space} & \textbf{Description} & \textbf{Best found}\\
\hline

$n_{\text{estimators}}$ & \{100, 200, 500, 800\} &
Number of trees in the forest. & $500$ \\

Criterion & \{gini, entropy\} &
Function used to measure the quality of a split. & entropy \\

$\text{min\_samples\_split}$ & \{2, 4, 6, 8, 10\} &
Minimum number of samples required to split an internal node. & $8$ \\

$\text{max\_features}$ & \{sqrt, log2, 0.4, 0.6, 0.8\} &
Number (or fraction) of features considered when looking for the best split. & sqrt \\

$\text{max\_depth}$ & \{None, 5, 10, 20, 30\} &
Maximum depth of the trees. & None \\

$\text{max\_samples}$ & Uniform(0.5, 1.0) &
Fraction of samples used to train each tree (bootstrap sampling). & $0.9206$ \\

$\text{min\_samples\_leaf}$ & \{1, 2, 5, 10\} &
Minimum number of samples required to be at a leaf node. & $5$ \\

\hline
\end{tabular}
\caption{Random search hyperparameter space and best-found hyperparameters for the Random Forest classifier. We used the Random Forest implementation from scikit-learn v.1.8.0.}
\label{tab:rf_random_search}
\end{table*}

\begin{table*}[h]
\centering
\begin{tabular}{llp{7cm}l}
\hline
\textbf{Hyperparameter} & \textbf{Search Space} & \textbf{Description} & \textbf{Best found} \\
\hline

$n_{\text{estimators}}$ & \{100, 200, 400, 800, 1200\} &
Number of boosting rounds (trees). & $800$ \\

Learning rate ($\eta$) & LogUniform($10^{-3}, 0.3$) &
Step size shrinkage used to prevent overfitting. &  $0.026$ \\

$\gamma$ & LogUniform($10^{-3}, 1.0$) &
Minimum loss reduction required to make a split (controls tree complexity). & $0.046$ \\

$\text{max\_depth}$ & \{3, 4, 5, 6, 7, 9\} &
Maximum depth of individual trees. & $6$ \\

$\text{min\_child\_weight}$ & \{1, 2, 4, 6, 8, 10, 12\} &
Minimum sum of instance weights needed in a child node. & $1$ \\

Subsample & Uniform(0.6, 1.0) &
Fraction of training samples used for each boosting round. & $0.940$ \\

$\text{colsample\_bytree}$ & Uniform(0.6, 1.0) &
Fraction of features sampled for each tree. & $0.732$ \\

$\alpha$ ($\text{reg\_alpha}$) & LogUniform($10^{-8}, 10$) &
L1 regularization term on weights. & $4.467 * 10^{-8}$\\

$\lambda$ ($\text{reg\_lambda}$) & LogUniform($10^{-2}, 100$) &
L2 regularization term on weights. & $0.347$ \\

\hline
\end{tabular}
\caption{Random search hyperparameter space and best-found hyperparameters for the XGBoost classifier. We used the XGBoost implementation from xgboost v.3.2.1.}
\label{tab:xgb_random_search}
\end{table*}

\begin{table*}[h!]
\centering
\begin{tabular}{lll}
\hline
\textbf{Hyperparameter} & \textbf{Search Space} & \textbf{Description} \\
\hline

Scaling type & \{min\_max, z\_score\} &
Input feature normalization method applied before training. \\

Activation (middle layers) & \{ReLU, Swish\} &
Activation function used in intermediate residual blocks. \\

Learning rate & \{ $10^{-3}$, $3 \times 10^{-3}$, $10^{-4}$ \} &
Optimizer step size controlling gradient updates. \\

Batch size & \{8, 16, 32, 48, 64\} &
Number of training samples processed before updating model weights. \\

\hline
\end{tabular}
\caption{Hyperparameter search space for the ResNet architecture. For both depths 18 and 34, the best hyperparameter values are z\_score, Swish, and $0.001$. For the batchsize, it is respectively $32$ and $16$. ResNet models were implemented using torch v.2.5.1.}
\label{tab:resnet_hyperparameter_search}
\end{table*}

\clearpage

\end{appendices}

\bibliographystyle{unsrt}
\bibliography{literature}
\end{document}